\documentclass[11pt]{article}

\usepackage[final]{acl}

\usepackage{times}
\usepackage{latexsym}
\usepackage{balance}

\usepackage[T1]{fontenc}

\usepackage[utf8]{inputenc}

\usepackage{microtype}

\usepackage{inconsolata}
\usepackage{multirow}
\usepackage{graphicx}
\usepackage{caption}
\usepackage{amsmath}
\usepackage{amssymb}
\usepackage{booktabs}
\usepackage{tikz}
\usepackage{enumitem}
\usepackage{setspace}
\usepackage[table]{xcolor}
\usepackage{float}
\usetikzlibrary{arrows.meta, positioning}
\title{When Evidence Conflicts: Reliability-aware Meta-review Generation}

\author{
  \textbf{Xinzhe Wang\textsuperscript{1}},
  \textbf{Fei Tao\textsuperscript{2}},
  \textbf{Jiang Xie\textsuperscript{1}},
  \textbf{Hong Yu\textsuperscript{1}},
  \textbf{Ye Wang\textsuperscript{1}}\thanks{Corresponding author: \texttt{wangye@cqupt.edu.cn}}
  \\
  \textsuperscript{1}  Chongqing University of Posts and Telecommunications, China
  \\
  \textsuperscript{2}NewsBreak, USA
}

\begin{document}
\maketitle

\begin{abstract}

Generating coherent meta-reviews from multiple peer reviews is challenging when reviewer evidence conflicts and varies in reliability. Existing approaches typically formulate meta-review generation as a multi-document summarization task and aggregate reviewer feedback uniformly, making it difficult to determine which opinions should be prioritized under disagreement. In this paper, we study meta-review generation through reliability-aware evidence aggregation. Our framework first extracts aspect-level opinions from peer reviews and identifies conflicting evidence within each aspect. It then estimates opinion-level support and review-level quality to measure evidence reliability. Based on these signals, the framework assigns reliability-aware weights to reviewer feedback, enabling the generator to prioritize better-supported arguments while preserving diverse perspectives. Experiments demonstrate that our method consistently improves meta-review generation over strong baselines on both automatic and human evaluations, with clear gains in conflict recognition and resolution under high-conflict review scenarios. The code and implementation details are publicly available at \url{https://github.com/Wangxz729/reliability-aware-meta-review}.

\end{abstract}

\section{Introduction}

Synthesizing multiple peer reviews into a meta-review is an important component of academic peer review. With the development of large language models (LLMs), recent work has explored meta-review generation by summarizing reviewer feedback from multiple sources \cite{du-etal-2024-llms,10.1007/978-981-97-9536-9_2,li-etal-2024-sentiment,hossain-etal-2025-llms, idahl-ahmadi-2025-openreviewer}. Despite promising progress, this task remains challenging when reviewer evidence conflicts. Reviewers may provide divergent assessments of the same paper in terms of novelty, soundness, empirical validity, or overall significance. Uniform aggregation can blur key disagreements and weaken the evidential basis of the meta-review.

Reviewer disagreement is common in academic peer review and often reflects differences in expertise, interpretation, or evaluation criteria rather than simple annotation noise \cite{basile-etal-2021-need,xu2026consensusperspectivistmodelingevaluation}. Prior work on disagreement modeling shows that conflicting judgments can offer complementary perspectives \cite{chu2021learning,basile-etal-2021-need,leonardelli-etal-2023-semeval}. Thus, meta-review generation should not simply collapse disagreement into a single summary. Instead, it should weigh conflicting evidence when forming the final meta-review.

A further challenge is that conflicting reviewer evidence is not equally reliable. Some comments provide concrete observations, detailed reasoning, and well-supported critiques, whereas others are vague, speculative, or weakly justified. Existing approaches uniformly aggregate reviewer comments, leading to difficulty in prioritizing opinions when evidence conflicts. Therefore, reviewer disagreement should not be resolved by majority sentiment or uniform summarization alone. The key question is which pieces of evidence deserve greater influence when forming the final meta-review. This view is related to argument quality assessment, which evaluates textual claims according to factors such as specificity, justification, and evidential support \cite{wachsmuth-etal-2017-computational,habernal-gurevych-2016-makes}.

Existing approaches provide limited mechanisms for addressing this problem. Most meta-review generation methods formulate the task as multi-document summarization and mainly improve generation quality through prompting strategies or iterative refinement \cite{li-etal-2023-summarizing,10.1007/978-981-97-9536-9_2,hossain-etal-2025-llms}. Sentiment-based consolidation methods can expose coarse agreement and disagreement patterns, but they typically treat opinions with the same polarity similarly \cite{li-etal-2024-sentiment}. More broadly, prior work on opinion and aspect-based summarization has explored structured opinion extraction, viewpoint consolidation, and evidence aggregation, but does not explicitly model the reliability of competing positive and negative opinions within the same review aspect. As a result, existing approaches can identify or organize reviewer disagreement, but still lack an explicit mechanism for determining which conflicting
evidence is better supported and should receive greater influence during generation.

To address this limitation, we investigate meta‑review generation via reliability‑aware evidence aggregation. Our framework first extracts aspect‑level opinions from peer reviews and detects conflicting evidence per aspect. It then estimates opinion‑level support and review‑level quality with an LLM‑based evaluator, following recent work on LLM-based evaluation \cite{NEURIPS2023_91f18a12}. Using these reliability signals, the framework assigns adaptive weights to reviewer evidence to prioritize better‑supported arguments during aggregation. Weighted evidence is structured into aspect‑level representations to guide the generator for final meta‑review production.

We evaluate our framework on the ORSum benchmark. Results show consistent improvements over strong baselines in both automatic and human evaluations, with larger gains on high-conflict cases. These findings highlight the value of reliability-aware aggregation when reviewer evidence conflicts.

Our contributions are summarized as follows:
\begin{itemize}
    \item We formulate meta-review generation under reviewer disagreement as a reliability-aware evidence aggregation problem, where conflicting opinions are modeled at the aspect level rather than at the whole-review
    level.

    \item We introduce a two-level reliability-aware aggregation framework that combines opinion-level support with review-content quality and uses these signals to conditionally down-weight weak evidence while preserving competing viewpoints.

    \item We provide empirical evidence through component ablations, evaluator robustness analysis, controlled weight/order experiments, human evaluation, and conflict-focused LLM evaluation on the ORSum benchmark.
\end{itemize}

\begin{figure*}[t]
    \centering
    \includegraphics[width=\textwidth]{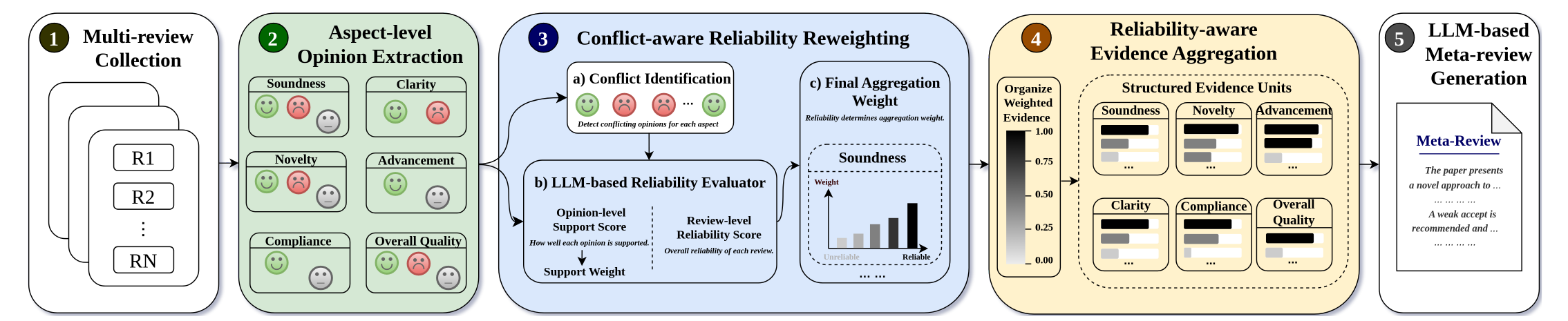}
    \caption{
        Overview of the proposed framework. The framework extracts aspect-level opinions from peer reviews, identifies conflicts within each aspect, estimates opinion-level support and review-content quality, performs conflict-conditioned reliability-aware evidence aggregation, and generates the final meta-review from the resulting weighted evidence.
    }
    \label{fig:framework}
\end{figure*}

\section{Related Work}



\subsection{Meta-review Generation}

Meta-review generation aims to synthesize multiple reviewer comments and discussions into a coherent meta-review for academic decision-making. Early studies formulate this task as a multi-document summarization problem over peer reviews and author-reviewer discussions \cite{li-etal-2023-summarizing,10.1007/978-981-97-9536-9_2}. Related work on multi-document summarization has also explored how to aggregate information from multiple source documents into coherent summaries \cite{fabbri-etal-2019-multi}.

More recently, meta-review generation has been studied in LLM-driven settings, where prompting strategies and multi-stage generation pipelines are commonly adopted. For example, CGI$^2$ decomposes meta-review generation into multiple steps with iterative self-refinement, while TELeR studies structured prompting strategies for assisting meta-review writing \cite{10.1007/978-981-97-9536-9_2,hossain-etal-2025-llms}. Other studies further examine the potential and risks of using LLMs in scholarly peer review and meta-reviewing \cite{du-etal-2024-llms,ye2024we}. Although these methods improve generation fluency and coverage, they largely treat reviewer evidence as uniformly reliable, leaving the problem of conflicting evidence underexplored.

\subsection{Evidence Aggregation in Opinion Summarization}

Meta-review generation can be viewed as an opinion aggregation problem, where reviewer comments provide evaluative evidence across different aspects. Prior work on opinion and argument-aware summarization has explored aspect extraction, sentiment analysis, viewpoint consolidation, and evidence-aware aggregation \cite{angelidis-etal-2021-extractive,wachsmuth-etal-2017-computational,habernal-gurevych-2016-makes,basile-etal-2021-need}. Recent studies further incorporate argumentation schemes, aspect key-point analysis, and decomposed aspect-aware modules for structured opinion summarization \cite{zhou-etal-2025-aspect,tang-etal-2024-prompted,li-etal-2025-decomposed}. However, scientific peer reviews pose a distinct challenge: opinions are aspect-dependent and argument-based, and positive and negative evaluations may directly conflict.

In peer-review summarization, recent sentiment-based consolidation methods use polarity signals to guide meta-review generation \cite{li-etal-2024-sentiment}. While such approaches make agreement and disagreement more explicit, sentiment alone does not indicate the quality or evidential strength of an opinion. Opinions with the same polarity may differ substantially in specificity, justification, and support. Our approach therefore focuses on estimating the relative reliability of competing opinions and incorporating both opinion-level support and review-level reliability into conflict-aware evidence aggregation.

\subsection{Reliability of Conflicting Evidence}

Conflicting evidence is common in subjective evaluation settings, where different judgments may reflect diverse perspectives rather than noise. Prior work on disagreement modeling shows that such disagreement can contain meaningful signals and should not always be reduced to a single consensus label \cite{basile-etal-2021-need,leonardelli-etal-2023-semeval,xu2026consensusperspectivistmodelingevaluation}. This is especially relevant to peer review, where reviewer disagreement may arise from differences in expertise, criteria, or interpretation.

Reliability has also been studied in crowdsourced learning and argument quality assessment. Crowdsourced learning estimates annotator reliability to reduce noisy supervision \cite{chu2021learning}, while argument quality assessment evaluates textual claims through specificity, justification, plausibility, and evidential support \cite{wachsmuth-etal-2017-computational,habernal-gurevych-2016-makes}. However, these reliability signals have not been fully used in meta-review generation under conflicting reviewer evidence. Our work addresses this gap by estimating opinion-level support and review-level quality, weighting reviewer evidence during generation.

\section{Method}

We propose a reliability-aware evidence aggregation framework for meta-review generation under reviewer disagreement. Instead of uniformly aggregating reviewer comments, our framework explicitly models the reliability of different opinions and prioritizes better-supported evidence during generation.

Figure~\ref{fig:framework} presents an overview of the proposed reliability-aware evidence aggregation framework. Starting from peer reviews, the framework extracts aspect-level opinions, identifies reviewer disagreement, estimates opinion reliability, and organizes reviewer evidence for meta-review generation.

Given a set of peer reviews for a paper, our goal is to generate a coherent meta-review that appropriately resolves conflicting reviewer opinions while preserving important supporting evidence.

In this paper, we distinguish between a \emph{review}, which refers to an entire reviewer report, and an \emph{opinion}, which denotes an aspect-level evaluative statement extracted from a review.

\subsection{Opinion Extraction and Conflict Characteristics}

We begin by extracting structured opinion units from raw peer reviews and organizing them by predefined review aspects, such as soundness, clarity, novelty, and overall quality. We perform aspect-level opinion extraction using an LLM-based structured extraction prompt. For each review, the model identifies aspect labels, opinion, and sentiment polarity labels (positive, neutral, negative). Extracted opinions are then normalized into structured opinion units for subsequent conflict analysis and aggregation. To improve consistency, all extraction prompts use deterministic decoding (temperature=0). Example extraction prompts are provided in Appendix~\ref{appendix:prompts}.

To better understand the nature of reviewer disagreement, we first analyze the distribution of sentiment across different aspects. Table~\ref{tab:sentiment_dist} presents the normalized sentiment distribution aggregated over the dataset.

\begin{table}[t]
\centering
\small
\renewcommand{\arraystretch}{1.2}
\setlength{\tabcolsep}{4pt}
\resizebox{\columnwidth}{!}{
\begin{tabular}{l c c c}
\hline
\noalign{\vspace{2pt}}
\textbf{Aspect} & \textbf{Positive (\%)} & \textbf{Neutral (\%)} & \textbf{Negative (\%)} \\
\noalign{\vspace{2pt}}
\hline
Soundness & 15.14 & 4.62 & 80.24 \\
Clarity & 31.96 & 6.75 & 61.29 \\
Advancement & 47.61 & 3.44 & 48.95 \\
Novelty & 59.52 & 4.56 & 35.92 \\
Overall Quality & 63.00 & 10.46 & 26.54 \\
Compliance & 11.51 & 32.85 & 55.64 \\
\hline
\end{tabular}
}
\caption{Sentiment distribution across review aspects (in percentage).}
\label{tab:sentiment_dist}
\end{table}

As shown in Table~\ref{tab:sentiment_dist}, different aspects exhibit substantially different sentiment patterns. For example, \emph{soundness} is dominated by negative opinions, indicating that reviewers tend to be more critical of methodological rigor. In contrast, \emph{novelty} and \emph{overall quality} receive relatively more positive feedback. Notably, aspects such as \emph{advancement} present a nearly balanced distribution between positive and negative sentiments, suggesting a higher likelihood of disagreement. This variation indicates that disagreement is not uniformly distributed, but depends on the nature of the evaluation aspect.

To further quantify the prevalence of disagreement, we compute the proportion of papers for which conflicting sentiments exist within each aspect. An aspect is considered conflicting if both positive and negative opinions are present. The results are shown in Table~\ref{tab:conflict_ratio}.

The results reveal that reviewer disagreement is both pervasive and highly aspect-dependent. In particular, aspects such as \emph{advancement}, \emph{clarity}, and \emph{novelty} exhibit high conflict ratios, indicating that reviewers frequently hold opposing views on these dimensions. In contrast, \emph{compliance} shows relatively low disagreement, suggesting more consistent and objective judgments.

To further characterize disagreement at the paper level,
we define the conflict ratio of a paper as the proportion
of aspects containing both positive and negative opinions
among all extracted aspects:

\begin{equation}
\mathrm{ConflictRatio}(p)=
\frac{
\sum_{a \in \mathcal{A}_p} C_a
}{
|\mathcal{A}_p|
},
\end{equation}

where $\mathcal{A}_p$ denotes the set of extracted aspects for paper $p$, and $C_a \in \{0,1\}$ indicates whether aspect $a$ contains conflicting opinions as defined in Equation~\ref{eq:conflict_indicator}.

These observations highlight two key challenges for meta-review generation. First, conflicts are widespread rather than exceptional, making naive aggregation strategies insufficient. Second, the degree of disagreement varies across aspects, implying that conflict handling should be adaptive rather than uniform. Moreover, aspects with higher conflict ratios tend to involve more subjective evaluation criteria, which further complicates reliable aggregation. These findings motivate the need for a conflict-aware mechanism that can explicitly account for disagreement and differentiate the influence of competing opinions.

\begin{table}[t]
\centering
\small
\renewcommand{\arraystretch}{1.2}
\setlength{\tabcolsep}{5pt}
\begin{tabular}{lc}
\hline
\noalign{\vspace{2pt}}
\textbf{Aspect} & \textbf{Conflict Ratio (\%)} \\
\noalign{\vspace{2pt}}
\hline
Soundness        & 42.05 \\
Clarity          & 59.56 \\
Advancement      & 64.52 \\
Novelty          & 53.94 \\
Overall Quality  & 40.44 \\
Compliance       & 7.95  \\
\hline
\end{tabular}
\caption{Proportion of papers with conflicting opinions per aspect.}
\label{tab:conflict_ratio}
\end{table}

\subsection{Conflict-aware Opinion Support and Down-weighting}
\label{sec:conflict}

To resolve reviewer disagreement in a principled manner, we propose a conflict-aware support estimation and down-weighting mechanism that assigns different importance to opinions based on their strength.
\paragraph{Conflict-aware Setting}
For each aspect $a$, we denote its associated opinion set as
\begin{equation}
\mathcal{O}_a = \{o_1, o_2, \dots, o_k\}.
\end{equation}
Each opinion $o \in \mathcal{O}_a$ is associated with a sentiment label $s(o) \in \{\mathrm{pos}, \mathrm{neg}, \mathrm{neu}\}$.
We define an aspect as conflicting only when both positive and negative opinions coexist under the same aspect:
\begin{equation}
C_a = \mathbb{I}\left[
\exists o_i,o_j \in \mathcal{O}_a:
s(o_i)=\mathrm{pos},
s(o_j)=\mathrm{neg}
\right]
\label{eq:conflict_indicator}
\end{equation}

﻿
For non-conflicting aspects, we retain all extracted opinions and apply review-level reliability through $q_r$. For conflicting aspects, we additionally apply opinion-level support weighting to differentiate the relative influence of competing positive and negative evidence.

\paragraph{Opinion Support Estimation}
To quantify the strength of each opinion, we evaluate four factors: \emph{specificity}, \emph{rationality}, \emph{justification}, and \emph{academic plausibility}. These dimensions are adapted from prior work on argument quality and convincingness \cite{wachsmuth-etal-2017-computational,habernal-gurevych-2016-makes, habernal-gurevych-2016-argument}. Each factor $f_i$ is rated on a discrete scale from $1$ to $5$ and normalized to the range $[0,1]$:
\begin{equation}
\tilde{f}_i = \frac{f_i - 1}{4}.
\end{equation}
The overall support score is computed as a weighted sum:
\begin{equation}
\mathrm{support}(o) = \sum_{i=1}^{4} \alpha_i \tilde{f}_i(o),
\end{equation}
where $\sum_{i=1}^{4} \alpha_i = 1$. In our implementation, we adopt uniform weights $\alpha_i = 0.25$ to avoid introducing additional task-specific hyperparameters and to prevent the scoring function from implicitly favoring one argument-quality dimension without sufficient empirical justification. We further examine the sensitivity of this choice in Appendix~\ref{app:weight_order}. The support score is estimated using an LLM-based evaluator according to predefined argument-quality criteria. Detailed prompting and scoring settings are described in Section~\ref{sec:exp}.

\paragraph{Down-weighting Mechanism}

Given a threshold $\tau \in [0,1]$, for opinions satisfying $\mathrm{support}(o) < \tau$, we compute a normalized support deficiency:

\begin{equation}
\delta(o)=
\frac{\tau-\mathrm{support}(o)}
{\max_{o' \in \mathcal{O}_a}
(\tau-\mathrm{support}(o'))}.
\end{equation}

We adopt a smooth reliability-aware gating function to determine the contribution of each opinion during aggregation.The purpose of this function is to reduce the influence of weakly supported evidence without completely discarding minority opinions, which may still contain important concerns. The resulting weight assigned to an opinion, denoted as the \emph{support weight}, is defined as:

\begin{equation}
w(o)=
\begin{cases}
1, & \mathrm{support}(o)\ge\tau, \\[4pt]
\sigma\bigl(\lambda (1-\delta(o))\bigr), & \mathrm{support}(o)<\tau.
\end{cases}
\end{equation}

where $\sigma(\cdot)$ denotes the sigmoid function, and $\lambda$ controls the sharpness of the weighting transition.

This formulation provides a smooth reliability-aware weighting strategy that progressively down-weights weakly supported opinions instead of discarding them entirely. We do not claim that the sigmoid form is theoretically optimal; rather, it is an empirically effective implementation of conflict-conditioned reliability weighting.

\paragraph{Discussion}
The proposed mechanism provides a smooth and continuous weighting strategy. Instead of discarding low-support opinions, it progressively reduces their influence while preserving diverse viewpoints. This design enables more robust and fine-grained resolution of conflicting reviewer opinions.

\begin{table*}[t]
\centering
\small
\setlength{\tabcolsep}{4pt}
\begin{tabular}{l|cccc|cccc}
\toprule
\multirow{2}{*}{Method} 
& \multicolumn{4}{c|}{Backbone Generator: Qwen3} 
& \multicolumn{4}{c}{Backbone Generator: DeepSeek} \\
\cmidrule(lr){2-5}
\cmidrule(lr){6-9}
& R-1 & R-2 & R-L & BERTScore
& R-1 & R-2 & R-L & BERTScore \\
\midrule

CGI$^2$ (24' IJCAI-W)
& 26.91 & 3.55 & 14.42 & 83.55
& 27.38 & 4.85 & 15.27 & 83.39 \\

TELeR Prompt level 3 (25' NAACL)
& 23.97 & 4.20 & 12.20 & 81.31
& 23.53 & 3.95 & 12.11 & 80.94 \\

TELeR Prompt level 4 (25' NAACL)
& 21.12 & \textbf{4.34} & 11.01 & 81.28
& 21.29 & 4.50 & 11.11 & 81.35 \\

Sentiment Consolidation (24' ACL)
& 22.57 & 4.28 & 11.33 & 82.00
& 24.78 & 4.75 & 13.07 & 82.61 \\

\midrule

Ours
& \textbf{27.90}$^\dagger$
& 4.17
& \textbf{14.54}
& \textbf{83.90}$^\dagger$
& \textbf{28.58}$^\dagger$
& \textbf{5.01}
& \textbf{15.29}
& \textbf{83.96}$^\dagger$ \\

\bottomrule
\end{tabular}
\caption{Main results on the ORSum benchmark. Our framework consistently improves ROUGE-1, ROUGE-L, and BERTScore across both backbone generators, while ROUGE-2 shows mixed results. $\dagger$ indicates statistically significant improvement over CGI$^2$ under the paired bootstrap test.}
\label{tab:main_results}
\end{table*}

\subsection{Review-level Reliability Modeling}

Beyond opinion-level support estimation, we further model review-level reliability to capture variation in review content quality across different reviews. The intuition is that some review texts provide clearer, more specific, and better-justified evidence than others. Ignoring such variation may lead to over-reliance on weakly supported review content even when individual opinions appear reasonable.

Importantly, our framework does not attempt to estimate the intrinsic competence or long-term credibility of individual reviewers. Instead, reliability estimation is applied only to the textual quality of review content within the current paper context.

To this end, we estimate a review-level reliability score based on several textual quality indicators, including clarity, specificity, constructiveness, and justification strength. Each review is associated with a reliability score $q_r \in [0,1]$, reflecting the estimated reliability of the review content for the current paper. Review-level reliability scores are also computed using an independent LLM-based evaluator following
predefined quality rubrics.

Formally, given an opinion $o$ extracted from review $r$, we incorporate review-level reliability by modulating its opinion-level support weight. The resulting \emph{final aggregation weight} is defined as:

\begin{equation}
\tilde{w}(o)=w(o)\cdot q_r,
\end{equation}

where $w(o)$ denotes the opinion-level support weight defined in Section~\ref{sec:conflict}, and $q_r$ represents the reliability score of review $r$.

This formulation ensures that opinions extracted from higher-quality review content contribute more strongly during aggregation, while still preserving diverse reviewer viewpoints. By combining opinion-level support estimation with review-level reliability modeling, the framework achieves more robust evidence aggregation under reviewer disagreement.

\subsection{Reliability-aware Evidence Aggregation}

Given a set of weighted opinions, we organize them into structured evidence units for meta-review generation. Each evidence unit contains the aspect label, opinion, sentiment polarity, and the final aggregation weight.

For each aspect $a$, we construct a ranked evidence set:

\begin{equation}
E_a = \mathrm{Sort}\left(
\{(o,\tilde{w}(o))\}_{o\in \mathcal{O}_a}
\right),
\end{equation}

where opinions are ordered by their final aggregation weights while the weights themselves are explicitly provided to the generator. The ordering is therefore used only to organize the evidence; the numerical weights constitute the primary reliability signal.

The final meta-review is generated by conditioning the language model on the aggregated evidence across all aspects:

\begin{equation}
M = G(\{E_a\}_{a\in \mathcal{A}}),
\end{equation}

where $G(\cdot)$ denotes the generator language model.

This generation method organizes evidence according to both aspect structure and estimated reliability. By explicitly exposing the numerical weights to the generator, the framework allows the generator to distinguish the relative influence of competing evidence rather than relying solely on their input order.

\section{Experiments}
\label{sec:exp}

\subsection{Experimental Setup}

\paragraph{Dataset and Baselines.}

We evaluate our framework on the ORSum benchmark \citep{10.1007/978-981-97-9536-9_2}, a widely used dataset for meta-review generation. ORSum contains peer reviews and corresponding meta-reviews written by area chairs across multiple scientific venues, making it particularly suitable for studying reviewer disagreement and opinion aggregation.

We compare our method with several representative baselines, including CGI$^2$ \citep{10.1007/978-981-97-9536-9_2}, TELeR \citep{hossain-etal-2025-llms}, and the Sentiment Consolidation Framework \citep{li-etal-2024-sentiment}.

\paragraph{Models and Evaluation Metrics.}

We conduct experiments using Qwen3 and DeepSeek as backbone generation models responsible for meta-review generation, while GPT-4o-mini is used separately as the evaluator model for reliability estimation. To assess the robustness of reliability estimation, we additionally evaluate GPT-4o-mini, Qwen3-32B, and DeepSeek-V3 as alternative evaluator models under a controlled setting.

Following prior work on meta-review generation and summarization, we adopt ROUGE-1, ROUGE-2, ROUGE-L, and BERTScore (F1) as automatic evaluation metrics. ROUGE measures lexical overlap between generated and reference meta-reviews, while BERTScore evaluates semantic similarity at the contextual embedding level.

To further analyze performance under reviewer disagreement, we additionally partition the test set into low-conflict, medium-conflict, and high-conflict subsets according to aspect-level conflict ratios.

\paragraph{Implementation Details.}

For reliability estimation, all opinion support scores and review-level reliability scores are computed using an LLM-based evaluator under deterministic decoding settings (temperature=0). The reliability threshold $\tau$ is set to 0.5 and the sigmoid sharpness parameter $\lambda$ is set to 5 in all experiments. Implementation details and hyperparameter settings are provided in Appendix~\ref{appendix:parameter}.

Reliability Estimation Details.
We use GPT-4o-mini as the evaluator model for both opinion support estimation and review-level reliability scoring. The evaluator is independent from the backbone generation models to reduce evaluation-generation coupling.

For each opinion, the evaluator scores four dimensions: specificity, rationality, justification strength, and academic plausibility. Each dimension is rated on a discrete 1--5 scale according to predefined rubrics. The final support score is computed as the normalized average across all dimensions.

For review-level reliability estimation, the evaluator assesses the overall quality of review content based on clarity, constructiveness, specificity, and evidential grounding. All evaluations are performed under deterministic decoding settings with temperature=0.

Complete scoring rubrics and prompts are provided in Appendix~\ref{appendix:rubric} and Appendix~\ref{appendix:prompts}.

\paragraph{Reliability Score Validation}

We randomly sample 200 opinions and have two annotators score opinion quality on specificity, rationality, justification, and academic plausibility. The estimated support scores correlate well with human judgments: Spearman 0.71, Pearson 0.68, and inter-annotator Cohen's $\kappa$ 0.74. This confirms that the LLM-based evaluator captures meaningful signals aligned with human perception of argument quality.

\subsection{Main Results}

Table~\ref{tab:main_results} compares performance across backbone models. Our framework consistently improves ROUGE-1 and BERTScore over the strongest baseline under both Qwen3 and DeepSeek. Compared with CGI$^2$, ROUGE-1 improves from 26.91 to 27.90 under Qwen3 and from 27.38 to 28.58 under DeepSeek, while BERTScore improves by 0.35 and 0.57 points, respectively. ROUGE-2 results are mixed under Qwen3, whereas our method achieves the highest ROUGE-2 under DeepSeek. We therefore avoid claiming uniform improvements across all automatic metrics.

Compared with CGI$^2$, which uses iterative refinement, our method further improves ROUGE and BERTScore by 1.5–2.0\%, indicating complementary gains from reliability-aware aggregation. TELeR, relying on prompts but ignoring evidence strength, performs worse on most metrics. Sentiment Consolidation treats opinions with the same polarity equally, while our framework prioritizes better-supported opinions, yielding more reliable meta-reviews.

Results are consistent across Qwen3 and DeepSeek, showing model-agnostic robustness. Paired bootstrap tests confirm statistically significant improvements on ROUGE-1 and BERTScore. Overall, the gains arise from reliability-aware evidence aggregation rather than generation variance.

\paragraph{Statistical significance.}
We perform 500 paired bootstrap resamples over the full ORSum test set. Compared with CGI$^2$, the strongest baseline on ROUGE-1 and BERTScore, the improvements are statistically significant on ROUGE-1 ($p < 0.01$) and BERTScore ($p < 0.05$).

\begin{table}[htb]
\centering
\small
\begin{tabular}{lcc}
\toprule
Evaluator LLM & R-1 & BERTScore \\
\midrule
GPT-4o-mini & 29.15 & 83.84 \\
Qwen3-32B & 28.96 & 83.81 \\
DeepSeek-V3 & 29.01 & 83.83 \\
\bottomrule
\end{tabular}
\caption{Robustness to the choice of reliability evaluator on 100
randomly sampled test papers. Downstream performance varies only
slightly across evaluator models.}
\label{tab:evaluator_robustness}
\end{table}

\subsection{Evaluator Robustness}
\label{sec:evaluator_robustness}

To examine whether the effectiveness of reliability estimation depends on
a particular evaluator model, we conduct a controlled sensitivity analysis
on 100 randomly sampled test papers. We use GPT-4o-mini, Qwen3-32B, and
DeepSeek-V3 as reliability evaluators while keeping the generator,
prompts, sampled papers, and decoding settings fixed. Only the evaluator
model is changed.

The maximum downstream variation is 0.19 ROUGE-1 and 0.03 BERTScore.
The evaluator models also show strong agreement in opinion-level
reliability estimation, with pairwise Spearman correlations of
0.82--0.87 and Kendall's $\tau$ values of 0.71--0.76. These results
suggest that the observed performance is not driven solely by one
particular evaluator model.

\begin{table}[htb]
\centering
\scriptsize
\setlength{\tabcolsep}{2.4pt}
\renewcommand{\arraystretch}{1.15}
\begin{tabular}{@{}lccc@{}}
\toprule
Method & Low (N=167) & Med (N=166) & High (N=167)\\
\midrule
CGI$^2$ (24' IJCAI-W) & 26.88 & 26.34 & 25.71 \\
TELeR L3 (25' NAACL)  & 23.51 & 22.98 & 22.83 \\
TELeR L4 (25' NAACL)  & 21.92 & 21.53 & 20.94 \\
Sent. Consol. (24' ACL) & 22.93 & 22.51 & 21.24 \\
\midrule
Ours & \textbf{27.82} & \textbf{27.83} & \textbf{28.02} \\
\bottomrule
\end{tabular}
\caption{ROUGE-1 results across different conflict levels. Our method achieves more prominent improvements in high-conflict scenarios, verifying the efficacy of reliability-aware aggregation for resolving reviewer disagreements.}
\label{tab:conflict_results}
\end{table}

\subsection{Performance Across Conflict Levels}

To assess our framework under reviewer disagreement, we evaluate methods on low-, medium-, and high-conflict subsets. Each subset contains 167, 166, and 167 papers, respectively.

Table~\ref{tab:conflict_results} shows that the performance gains of our method increase with conflict level. On the high-conflict subset, our framework achieves ROUGE-1 of 28.02, outperforming CGI$^2$, which scores 25.71, and TELeR L4, which scores 20.94. The absolute ROUGE-1 scores are also higher in high-conflict cases, likely due to richer evaluative evidence. The key observation is the substantially larger relative gains achieved by our framework under severe disagreement.

On the low-conflict subset, our framework achieves ROUGE-1 of 27.82, compared to 26.88 for CGI$^2$, showing stable improvements even when reviewer agreement is high. These results support our central hypothesis that explicitly modeling evidence reliability is crucial for resolving conflicting reviewer opinions.

\subsection{Ablation Study}

We conduct ablation studies to evaluate the contribution of each component of our framework (Table~\ref{tab:ablation}): 
\textbf{Uniform} assigns equal weights to all opinions; 
\textbf{Support-only} uses only opinion-level support; 
\textbf{Reliability-only} uses only review-level reliability; 
\textbf{No Conflict Modeling} ranks opinions by reliability but ignores explicit conflict-aware grouping; 
\textbf{Full Model} denotes the complete framework.

Removing review-level reliability reduces ROUGE-1 from 27.90 to 27.35 and BERTScore from 83.90 to 83.61, while removing opinion-level support drops ROUGE-1 to 23.08 and BERTScore to 81.97. Eliminating conflict-aware aggregation slightly reduces ROUGE-1 to 24.32 and BERTScore to 82.38, especially on high-conflict samples. These results indicate that opinion-level support contributes most, but all components are complementary; the full model consistently achieves the best performance.

\begin{table}[htb]
\centering
\small
\setlength{\tabcolsep}{5pt}
\begin{tabular}{lcc}
\toprule
Variant
& R-1
& BERTScore \\
\midrule

Uniform
& 22.71
& 81.22 \\

Support-only
& 27.35
& 83.61 \\

Reliability-only
& 23.08
& 81.97 \\

No Conflict Modeling
& 24.32
& 82.38 \\

\midrule

Full Model
& \textbf{27.90}
& \textbf{83.90} \\

\bottomrule
\end{tabular}
\caption{
Ablation study of different framework components.
Both opinion-level support estimation and review-level reliability contribute substantially to overall performance.
}
\label{tab:ablation}
\end{table}

\begin{table*}[h]
\centering
\small
\setlength{\tabcolsep}{6pt}
\begin{tabular}{lcccc}
\toprule
Method
& Conflict Awareness
& Resolution Quality
& Evidence Grounding
& Coherence \\
\midrule

CGI$^2$ (24' IJCAI-W)
& 2.81
& 2.74
& 3.69
& 3.92 \\

TELeR Prompt level 3 (25' NAACL)
& 2.57
& 2.78
& 3.41
& 3.71 \\

TELeR Prompt level 4 (25' NAACL)
& 2.43
& 2.51
& 3.68
& 3.62 \\

Sentiment Consolidation (24' ACL)
& 3.64
& 3.08
& 3.79
& 3.65 \\

\midrule

Ours
& \textbf{4.56}
& \textbf{4.21}
& \textbf{4.18}
& \textbf{3.94} \\

\bottomrule
\end{tabular}
\caption{
Human evaluation results on a 1--5 Likert scale.
Our framework achieves the highest ratings across all dimensions, particularly on conflict awareness and resolution quality.
}
\label{tab:human_eval}
\end{table*}

\subsection{Human Evaluation}

Automatic metrics may not fully capture a system's ability to recognize and resolve reviewer disagreement. We therefore conduct human evaluation focusing on conflict handling quality.

We randomly sample 104 papers from the test set and evaluate generated meta-reviews through human assessment. Two independent annotators conduct blind evaluation using a 1--5 Likert scale across the following dimensions (see Appendix~\ref{app:human_eval} for full human evaluation protocol).:

\begin{itemize}
    \item \textbf{Conflict Awareness}: whether the generated meta-review correctly recognizes reviewer disagreement.
    
    \item \textbf{Resolution Quality}: whether conflicting opinions are resolved coherently and reasonably.
    
    \item \textbf{Evidence Grounding}: whether the generated conclusions are supported by reviewer evidence.
    
    \item \textbf{Overall Coherence}: the overall readability, fluency, and consistency of the generated meta-review.
\end{itemize}

The average Cohen’s $\kappa$ between annotators is 0.72, indicating substantial agreement.

As shown in Table~\ref{tab:human_eval}, our framework achieves the highest ratings across all dimensions. The largest improvements appear in Conflict Awareness and Resolution Quality, confirming the effectiveness of reliability-aware evidence aggregation under reviewer disagreement. Compared with sentiment-based aggregation, our method produces meta-reviews that are better grounded and less prone to overemphasizing unsupported opinions.

\begin{table}[t]
\centering
\small
\setlength{\tabcolsep}{3.5pt}
\begin{tabular}{p{2.45cm}ccc}
\toprule
Method & Conflict & Resolution & Evidence \\
\midrule
CGI$^2$ & 3.02 & 2.54 & \textbf{3.92} \\
TELeR (Level 3) & 2.23 & 2.57 & 3.56 \\
TELeR (Level 4) & 2.45 & 2.48 & 3.62 \\
Senti. Consol. & 3.56 & 3.15 & 3.68 \\
Ours & \textbf{4.31} & \textbf{3.83} & 3.85 \\
\bottomrule
\end{tabular}
\caption{GPT-4o-based conflict-focused evaluation on the same 104 papers
used for human evaluation. The judge assesses the generated meta-review
against the source reviews without access to the reference meta-review.}
\label{tab:llm_judge}
\end{table}

\subsection{Conflict-focused LLM Evaluation}
\label{sec:llm_judge}

Automatic reference-based metrics may not fully capture whether a generated meta-review recognizes disagreement, resolves competing opinions, and grounds its conclusions in the source reviews. We therefore conduct a complementary LLM-as-a-judge evaluation using GPT-4o on the same 104 papers used for human evaluation.

GPT-4o is given the source reviews and one generated meta-review, with system identities and the reference meta-review hidden. The judge evaluates each output independently using a fixed 1--5 rubric covering Conflict Awareness, Resolution Quality, and Evidence Grounding. We use deterministic decoding and the same evaluation prompt for all systems.

As shown in Table~\ref{tab:llm_judge}, our framework achieves the highest scores on all three dimensions, with the largest gains in Conflict Awareness and Resolution Quality. This suggests that its improvements extend beyond reference-based similarity metrics. A concrete case study illustrating conflict‑aware aggregation is presented in Appendix~\ref{app:case_study}.

\section{Conclusion}

Meta-review generation is challenging when reviewer opinions conflict, as uniform aggregation can obscure important disagreements and weaken evidence. To address this, we propose a reliability-aware evidence aggregation framework that explicitly evaluates the quality of each opinion and review. Our method combines opinion-level argument support, review-level quality signals, and conflict-aware weighting to prioritize better-supported evidence during meta-review generation. Experiments on the ORSum benchmark demonstrate consistent improvements over strong baselines in both automatic metrics and human evaluations, with particularly larger gains on high-conflict cases. These results provide empirical evidence that explicitly modeling evidence reliability can improve meta-review generation.

More broadly, our work highlights the potential of reliability-aware evidence aggregation for scientific opinion summarization and provides a basis for further study of trustworthy, LLM-assisted peer review systems.

\section*{Limitations}

While our framework improves meta-review generation under reviewer disagreement, several limitations remain. First, reliability estimation depends on LLM-based evaluation, which may introduce biases or inconsistencies. Second, our evaluation is conducted on the ORSum benchmark alone, cross-dataset validation is an important direction for future work. Finally, minority opinions may sometimes highlight important flaws, and reliability-aware aggregation could underemphasize these dissenting viewpoints. 

Future work could explore confidence-aware aggregation, richer conflict modeling, and human-in-the-loop meta-review generation.

\section*{Ethical Considerations}

Our framework is intended to assist, not replace, human reviewers. Reliability scores assess review content rather than reviewer competence, and final decisions should remain with humans.

\section*{Acknowledgments}

This work was supported by the National Natural Science Foundation of China (62306056, 62136002, and 62221005).

\bibliography{custom}

@inproceedings{li-etal-2023-summarizing,
    title = "Summarizing Multiple Documents with Conversational Structure for Meta-Review Generation",
    author = "Li, Miao  and
      Hovy, Eduard  and
      Lau, Jey",
    editor = "Bouamor, Houda  and
      Pino, Juan  and
      Bali, Kalika",
    booktitle = "Findings of the Association for Computational Linguistics: EMNLP 2023",
    month = dec,
    year = "2023",
    address = "Singapore",
    publisher = "Association for Computational Linguistics",
    url = "https://aclanthology.org/2023.findings-emnlp.472/",
    doi = "10.18653/v1/2023.findings-emnlp.472",
    pages = "7089--7112"
}

@inproceedings{hossain-etal-2025-llms,
    title = "{LLM}s as Meta-Reviewers' Assistants: A Case Study",
    author = "Hossain, Eftekhar  and
      Sinha, Sanjeev Kumar  and
      Bansal, Naman  and
      Knipper, Alex  and
      Sarkar, Souvika  and
      Salvador, John  and
      Mahajan, Yash  and
      Guttikonda, Sri  and
      Akter, Mousumi  and
      Hassan, Md. Mahadi  and
      Freestone, Matthew  and
      Williams Jr., Matthew C.  and
      Feng, Dongji  and
      Karmaker, Santu",
    editor = "Chiruzzo, Luis  and
      Ritter, Alan  and
      Wang, Lu",
    booktitle = "Proceedings of the 2025 Conference of the Nations of the Americas Chapter of the Association for Computational Linguistics: Human Language Technologies (Volume 1: Long Papers)",
    month = apr,
    year = "2025",
    address = "Albuquerque, New Mexico",
    publisher = "Association for Computational Linguistics",
    url = "https://aclanthology.org/2025.naacl-long.395/",
    doi = "10.18653/v1/2025.naacl-long.395",
    pages = "7763--7803",
    ISBN = "979-8-89176-189-6"
}

@inproceedings{li-etal-2024-sentiment,
    title = "A Sentiment Consolidation Framework for Meta-Review Generation",
    author = "Li, Miao  and
      Lau, Jey Han  and
      Hovy, Eduard",
    editor = "Ku, Lun-Wei  and
      Martins, Andre  and
      Srikumar, Vivek",
    booktitle = "Proceedings of the 62nd Annual Meeting of the Association for Computational Linguistics (Volume 1: Long Papers)",
    month = aug,
    year = "2024",
    address = "Bangkok, Thailand",
    publisher = "Association for Computational Linguistics",
    url = "https://aclanthology.org/2024.acl-long.547/",
    doi = "10.18653/v1/2024.acl-long.547",
    pages = "10158--10177"
}

@InProceedings{10.1007/978-981-97-9536-9_2,
    author="Zeng, Qi
    and Sidhu, Mankeerat
    and Blume, Ansel
    and Chan, Hou Pong
    and Wang, Lu
    and Ji, Heng",
    editor="Yin, Wenpeng
    and Ahn, Jihyun Janice
    and Zhang, Rui
    and Huang, Lifu
    and Hadfi, Rafik
    and Ito, Takayuki
    and Ohnuma, Susumu
    and Shiramatsu, Shun",
    title="Scientific Opinion Summarization: Paper Meta-review Generation Dataset, Methods, and Evaluation",
    booktitle="Artificial Intelligence for Research and Democracy",
    year="2025",
    publisher="Springer Nature Singapore",
    address="Singapore",
    pages="20--38",
    isbn="978-981-97-9536-9"
}

@inproceedings{chu2021learning,
  title={Learning from crowds by modeling common confusions},
  author={Chu, Zhendong and Ma, Jing and Wang, Hongning},
  booktitle={Proceedings of the AAAI Conference on Artificial Intelligence},
  volume={35},
  pages={5832--5840},
  year={2021}
}

@misc{xu2026consensusperspectivistmodelingevaluation,
      title={Beyond Consensus: Perspectivist Modeling and Evaluation of Annotator Disagreement in NLP}, 
      author={Yinuo Xu and David Jurgens},
      year={2026},
      eprint={2601.09065},
      archivePrefix={arXiv},
      primaryClass={cs.CL},
      url={https://arxiv.org/abs/2601.09065}, 
}

@inproceedings{du-etal-2024-llms,
    title = "{LLM}s Assist {NLP} Researchers: Critique Paper (Meta-)Reviewing",
    author = "Du, Jiangshu  and
      Wang, Yibo  and
      Zhao, Wenting  and
      Deng, Zhongfen  and
      Liu, Shuaiqi  and
      Lou, Renze  and
      Zou, Henry Peng  and
      Venkit, Pranav Narayanan  and
      Zhang, Nan  and
      Srinath, Mukund  and
      Zhang, Haoran Ranran  and
      Gupta, Vipul  and
      Li, Yinghui  and
      Li, Tao  and
      Wang, Fei  and
      Liu, Qin  and
      Liu, Tianlin  and
      Gao, Pengzhi  and
      Xia, Congying  and
      Xing, Chen  and
      Jiayang, Cheng  and
      Wang, Zhaowei  and
      Su, Ying  and
      Shah, Raj Sanjay  and
      Guo, Ruohao  and
      Gu, Jing  and
      Li, Haoran  and
      Wei, Kangda  and
      Wang, Zihao  and
      Cheng, Lu  and
      Ranathunga, Surangika  and
      Fang, Meng  and
      Fu, Jie  and
      Liu, Fei  and
      Huang, Ruihong  and
      Blanco, Eduardo  and
      Cao, Yixin  and
      Zhang, Rui  and
      Yu, Philip S.  and
      Yin, Wenpeng",
    editor = "Al-Onaizan, Yaser  and
      Bansal, Mohit  and
      Chen, Yun-Nung",
    booktitle = "Proceedings of the 2024 Conference on Empirical Methods in Natural Language Processing",
    month = nov,
    year = "2024",
    address = "Miami, Florida, USA",
    publisher = "Association for Computational Linguistics",
    url = "https://aclanthology.org/2024.emnlp-main.292/",
    doi = "10.18653/v1/2024.emnlp-main.292",
    pages = "5081--5099"
}

@inproceedings{basile-etal-2021-need,
    title = "We Need to Consider Disagreement in Evaluation",
    author = "Basile, Valerio  and
      Fell, Michael  and
      Fornaciari, Tommaso  and
      Hovy, Dirk  and
      Paun, Silviu  and
      Plank, Barbara  and
      Poesio, Massimo  and
      Uma, Alexandra",
    editor = "Church, Kenneth  and
      Liberman, Mark  and
      Kordoni, Valia",
    booktitle = "Proceedings of the 1st Workshop on Benchmarking: Past, Present and Future",
    month = aug,
    year = "2021",
    address = "Online",
    publisher = "Association for Computational Linguistics",
    url = "https://aclanthology.org/2021.bppf-1.3/",
    doi = "10.18653/v1/2021.bppf-1.3",
    pages = "15--21"
}

@inproceedings{wachsmuth-etal-2017-computational,
    title = "Computational Argumentation Quality Assessment in Natural Language",
    author = "Wachsmuth, Henning  and
      Naderi, Nona  and
      Hou, Yufang  and
      Bilu, Yonatan  and
      Prabhakaran, Vinodkumar  and
      Thijm, Tim Alberdingk  and
      Hirst, Graeme  and
      Stein, Benno",
    editor = "Lapata, Mirella  and
      Blunsom, Phil  and
      Koller, Alexander",
    booktitle = "Proceedings of the 15th Conference of the {E}uropean Chapter of the Association for Computational Linguistics: Volume 1, Long Papers",
    month = apr,
    year = "2017",
    address = "Valencia, Spain",
    publisher = "Association for Computational Linguistics",
    url = "https://aclanthology.org/E17-1017/",
    pages = "176--187"
}

@inproceedings{NEURIPS2023_91f18a12,
 author = {Zheng, Lianmin and Chiang, Wei-Lin and Sheng, Ying and Zhuang, Siyuan and Wu, Zhanghao and Zhuang, Yonghao and Lin, Zi and Li, Zhuohan and Li, Dacheng and Xing, Eric and Zhang, Hao and Gonzalez, Joseph and Stoica, Ion},
 booktitle = {Advances in Neural Information Processing Systems},
 editor = {A. Oh and T. Naumann and A. Globerson and K. Saenko and M. Hardt and S. Levine},
 pages = {46595--46623},
 publisher = {Curran Associates, Inc.},
 title = {Judging LLM-as-a-Judge with MT-Bench and Chatbot Arena},
 url = {https://proceedings.neurips.cc/paper_files/paper/2023/file/91f18a1287b398d378ef22505bf41832-Paper-Datasets_and_Benchmarks.pdf},
 volume = {36},
 year = {2023}
}

@inproceedings{habernal-gurevych-2016-makes,
    title = "What makes a convincing argument? Empirical analysis and detecting attributes of convincingness in Web argumentation",
    author = "Habernal, Ivan  and
      Gurevych, Iryna",
    editor = "Su, Jian  and
      Duh, Kevin  and
      Carreras, Xavier",
    booktitle = "Proceedings of the 2016 Conference on Empirical Methods in Natural Language Processing",
    month = nov,
    year = "2016",
    address = "Austin, Texas",
    publisher = "Association for Computational Linguistics",
    url = "https://aclanthology.org/D16-1129/",
    doi = "10.18653/v1/D16-1129",
    pages = "1214--1223"
}

@inproceedings{fabbri-etal-2019-multi,
    title = "Multi-News: A Large-Scale Multi-Document Summarization Dataset and Abstractive Hierarchical Model",
    author = "Fabbri, Alexander  and
      Li, Irene  and
      She, Tianwei  and
      Li, Suyi  and
      Radev, Dragomir",
    editor = "Korhonen, Anna  and
      Traum, David  and
      M{\`a}rquez, Llu{\'i}s",
    booktitle = "Proceedings of the 57th Annual Meeting of the Association for Computational Linguistics",
    month = jul,
    year = "2019",
    address = "Florence, Italy",
    publisher = "Association for Computational Linguistics",
    url = "https://aclanthology.org/P19-1102/",
    doi = "10.18653/v1/P19-1102",
    pages = "1074--1084"
}

@article{angelidis-etal-2021-extractive,
    title = "Extractive Opinion Summarization in Quantized Transformer Spaces",
    author = "Angelidis, Stefanos  and
      Amplayo, Reinald Kim  and
      Suhara, Yoshihiko  and
      Wang, Xiaolan  and
      Lapata, Mirella",
    editor = "Roark, Brian  and
      Nenkova, Ani",
    journal = "Transactions of the Association for Computational Linguistics",
    volume = "9",
    year = "2021",
    address = "Cambridge, MA",
    publisher = "MIT Press",
    url = "https://aclanthology.org/2021.tacl-1.17/",
    doi = "10.1162/tacl_a_00366",
    pages = "277--293"
}

@article{ye2024we,
  title={Are we there yet? revealing the risks of utilizing large language models in scholarly peer review},
  author={Ye, Rui and Pang, Xianghe and Chai, Jingyi and Chen, Jiaao and Yin, Zhenfei and Xiang, Zhen and Dong, Xiaowen and Shao, Jing and Chen, Siheng},
  journal={arXiv preprint arXiv:2412.01708},
  year={2024}
}

@inproceedings{leonardelli-etal-2023-semeval,
    title = "{S}em{E}val-2023 Task 11: Learning with Disagreements ({L}e{W}i{D}i)",
    author = "Leonardelli, Elisa  and
      Abercrombie, Gavin  and
      Almanea, Dina  and
      Basile, Valerio  and
      Fornaciari, Tommaso  and
      Plank, Barbara  and
      Rieser, Verena  and
      Uma, Alexandra  and
      Poesio, Massimo",
    editor = {Ojha, Atul Kr.  and
      Do{\u{g}}ru{\"o}z, A. Seza  and
      Da San Martino, Giovanni  and
      Tayyar Madabushi, Harish  and
      Kumar, Ritesh  and
      Sartori, Elisa},
    booktitle = "Proceedings of the 17th International Workshop on Semantic Evaluation (SemEval-2023)",
    month = jul,
    year = "2023",
    address = "Toronto, Canada",
    publisher = "Association for Computational Linguistics",
    url = "https://aclanthology.org/2023.semeval-1.314/",
    doi = "10.18653/v1/2023.semeval-1.314",
    pages = "2304--2318"
}

@inproceedings{idahl-ahmadi-2025-openreviewer,
    title = "{O}pen{R}eviewer: A Specialized Large Language Model for Generating Critical Scientific Paper Reviews",
    author = "Idahl, Maximilian  and
      Ahmadi, Zahra",
    editor = "Dziri, Nouha  and
      Ren, Sean (Xiang)  and
      Diao, Shizhe",
    booktitle = "Proceedings of the 2025 Conference of the Nations of the Americas Chapter of the Association for Computational Linguistics: Human Language Technologies (System Demonstrations)",
    month = apr,
    year = "2025",
    address = "Albuquerque, New Mexico",
    publisher = "Association for Computational Linguistics",
    url = "https://aclanthology.org/2025.naacl-demo.44/",
    doi = "10.18653/v1/2025.naacl-demo.44",
    pages = "550--562",
    ISBN = "979-8-89176-191-9"
}

@inproceedings{zhou-etal-2025-aspect,
    title = "Aspect-Based Opinion Summarization with Argumentation Schemes",
    author = {Zhou, Wendi  and
      Saadat-Yazdi, Ameer  and
      K{\"o}kciyan, Nadin},
    editor = "Chistova, Elena  and
      Cimiano, Philipp  and
      Haddadan, Shohreh  and
      Lapesa, Gabriella  and
      Ruiz-Dolz, Ramon",
    booktitle = "Proceedings of the 12th Argument mining Workshop",
    month = jul,
    year = "2025",
    address = "Vienna, Austria",
    publisher = "Association for Computational Linguistics",
    url = "https://aclanthology.org/2025.argmining-1.11/",
    doi = "10.18653/v1/2025.argmining-1.11",
    pages = "116--125",
    ISBN = "979-8-89176-258-9"
}

@inproceedings{tang-etal-2024-prompted,
    title = "Prompted Aspect Key Point Analysis for Quantitative Review Summarization",
    author = "Tang, An Quang  and
      Zhang, Xiuzhen  and
      Dinh, Minh Ngoc  and
      Cambria, Erik",
    editor = "Ku, Lun-Wei  and
      Martins, Andre  and
      Srikumar, Vivek",
    booktitle = "Proceedings of the 62nd Annual Meeting of the Association for Computational Linguistics (Volume 1: Long Papers)",
    month = aug,
    year = "2024",
    address = "Bangkok, Thailand",
    publisher = "Association for Computational Linguistics",
    url = "https://aclanthology.org/2024.acl-long.576/",
    doi = "10.18653/v1/2024.acl-long.576",
    pages = "10691--10708"
}

@inproceedings{li-etal-2025-decomposed,
    title = "Decomposed Opinion Summarization with Verified Aspect-Aware Modules",
    author = "Li, Miao  and
      Lau, Jey Han  and
      Hovy, Eduard  and
      Lapata, Mirella",
    editor = "Che, Wanxiang  and
      Nabende, Joyce  and
      Shutova, Ekaterina  and
      Pilehvar, Mohammad Taher",
    booktitle = "Findings of the Association for Computational Linguistics: ACL 2025",
    month = jul,
    year = "2025",
    address = "Vienna, Austria",
    publisher = "Association for Computational Linguistics",
    url = "https://aclanthology.org/2025.findings-acl.1273/",
    doi = "10.18653/v1/2025.findings-acl.1273",
    pages = "24805--24841",
    ISBN = "979-8-89176-256-5"
}

@inproceedings{habernal-gurevych-2016-argument,
    title = "Which argument is more convincing? Analyzing and predicting convincingness of Web arguments using bidirectional {LSTM}",
    author = "Habernal, Ivan  and
      Gurevych, Iryna",
    editor = "Erk, Katrin  and
      Smith, Noah A.",
    booktitle = "Proceedings of the 54th Annual Meeting of the Association for Computational Linguistics (Volume 1: Long Papers)",
    month = aug,
    year = "2016",
    address = "Berlin, Germany",
    publisher = "Association for Computational Linguistics",
    url = "https://aclanthology.org/P16-1150/",
    doi = "10.18653/v1/P16-1150",
    pages = "1589--1599"
}

\newpage
\appendix

\section{Weight and Ordering Analysis}
\label{app:weight_order}

\subsection{Controlled analysis of numerical weights and evidence ordering.}

To distinguish the effect of explicit numerical weights from the effect of evidence ordering, we conduct a controlled experiment on 100 randomly sampled test papers. All settings use the same sampled papers, generator, prompts, and decoding configuration; only the numerical-weight information and evidence order are changed.

Under random ordering, adding explicit numerical weights improves ROUGE-1 by 1.12 and BERTScore by 0.24. With numerical weights retained, sorting the evidence adds 0.21 ROUGE-1 and 0.02 BERTScore. These results suggest that numerical weights provide the primary contribution, whereas evidence ordering has a smaller complementary effect. Thus, our approach does not rely solely on placing high-weight evidence earlier in the input.

\begin{table}[htb]
\centering
\small
\begin{tabular}{lcc}
\toprule
Setting & ROUGE-1 & BERTScore \\
\midrule
Weight + descending order & 29.15 & 83.84 \\
Weight + random order & 28.94 & 83.82 \\
No weight + random order & 27.82 & 83.58 \\
\bottomrule
\end{tabular}
\caption{Controlled analysis of numerical weights and evidence ordering. Explicit numerical weights provide the primary gain, while ordering provides a smaller complementary effect.}
\label{tab:weight_order}
\end{table}

\subsection{Sensitivity to Support-Factor Weights}
\label{app:weight_sensitivity}

We further examine whether the equal weighting of the four support dimensions materially affects performance. We compare the default equal weighting with moderate changes that emphasize justification or specificity.

The maximum variation is 0.09 ROUGE-1 and 0.03 BERTScore, indicating limited sensitivity to moderate changes in factor weighting. This supports our use of equal weights as a simple setting that avoids introducing additional task-specific hyperparameters.

\begin{table}[htb]
\centering
\small
\begin{tabular}{lcc}
\toprule
Weighting Scheme & ROUGE-1 & BERTScore \\
\midrule
Equal $(1,1,1,1)$ & 29.15 & 83.84 \\
Justification-heavy $(1,1,2,1)$ & 29.17 & 83.84 \\
Specificity-heavy $(2,1,1,1)$ & 29.08 & 83.81 \\
\bottomrule
\end{tabular}
\caption{Sensitivity to the relative weighting of the four opinion-support dimensions. Moderate changes produce only small performance differences.}
\label{tab:weight_sensitivity}
\end{table}

\section{Hyperparameter Settings}
\label{appendix:parameter}

Table~\ref{tab:hyperparams} summarizes the main hyperparameter settings used in our experiments.

\begin{table}[H]
\centering
\small
\begin{tabular}{ll}
\toprule
Parameter & Value \\
\midrule
Support threshold $\tau$ & 0.5 \\
Evaluator temperature & 0 \\
Generator temperature & 0.7 \\
Max generation length & 512 \\
Top-p & 0.9 \\
$\lambda$ & 5 \\
\bottomrule
\end{tabular}
\caption{Main hyperparameter settings.}
\label{tab:hyperparams}
\end{table}

\section{Reliability Scoring Rubric}
\label{appendix:rubric}

Table~\ref{tab:scoring_rubric} summarizes the scoring criteria
used for opinion-level support factor evaluation.
The rubric is designed to encourage discriminative scoring
across opinions with different levels of evidence quality.
Higher scores indicate that an opinion is more concrete,
better justified, and more academically reliable, thereby
receiving a larger support weight during aggregation.

\begin{table}[htb]
\centering
\footnotesize
\setlength{\tabcolsep}{3pt}
\begin{tabular}{c >{\raggedright\arraybackslash}p{1.35cm} >{\raggedright\arraybackslash}p{1.55cm} >{\raggedright\arraybackslash}p{1.65cm} >{\raggedright\arraybackslash}p{1.65cm}}
\toprule
Score & Specificity & Rationality & Justification & Academic Plausibility \\
\midrule
1 & Vague or generic & Illogical or emotional & No supporting evidence & Academically weak \\
3 & Partially concrete & Reasonably logical & Some supporting rationale & Reasonable judgment \\
5 & Highly concrete & Well-reasoned and consistent & Strong supporting evidence & Academically convincing \\
\bottomrule
\end{tabular}
\caption{Scoring rubric for opinion-level support factor evaluation.}
\label{tab:scoring_rubric}
\end{table}

Table~\ref{tab:review_quality_rubric} presents the review-level
quality dimensions used for reviewer reliability estimation.
These dimensions evaluate the overall quality of reviewer comments
from multiple perspectives, including clarity, specificity,
constructiveness, and professionalism.

\begin{table}[H]
\centering
\scriptsize
\setlength{\tabcolsep}{3pt}

\begin{tabular}{p{2.2cm}p{4.8cm}}
\toprule
Dimension & Description \\
\midrule
Clarity &
Whether the review comments are easy to understand and clearly written.
\\
Specificity &
Whether the review discusses concrete aspects of the paper.
\\
Constructiveness &
Whether the review provides actionable suggestions or useful feedback.
\\
Professionalism &
Whether the review maintains an objective and professional tone.
\\
\bottomrule
\end{tabular}

\caption{Review-level quality dimensions used for reliability estimation.}
\label{tab:review_quality_rubric}

\end{table}

\section{Human Evaluation Protocol}
\label{app:human_eval}

Both annotators were graduate NLP researchers who had previously served as reviewers for peer-reviewed NLP/AI conferences or journals. The evaluation was conducted independently and blindly, with system identities hidden from the annotators. For each paper, the final score for each dimension was obtained by averaging the two annotators' ratings without an additional adjudication step. The mean Cohen's $\kappa$ across the four evaluation dimensions was 0.72.

\section{Case Study}
\label{app:case_study}

To further illustrate the effectiveness of the proposed reliability-aware aggregation framework, we present an additional case study from the ORSum benchmark. 

Table~\ref{tab:case_study} shows an example where reviewers express conflicting opinions regarding the novelty of the submission. While multiple reviewers question the originality of the proposed method, other reviewers emphasize the empirical effectiveness and robustness improvements. Our framework successfully identifies the disagreement and generates a balanced meta-review that reflects both supportive and critical viewpoints.

\begin{table}[htb]
\centering
\small
\setlength{\tabcolsep}{3pt}
\begin{tabular}{p{1.2cm} p{6.0cm}}
\toprule
Aspect & Novelty \\
\midrule

R1 &
``...The paper proposes a method for learning robust binary neural networks from random initialization. ...'' (Positive) \\

R2 &
``...The paper's proposed three techniques are not systematical but auxiliary model compression strategies. ...'' (Negative) \\

R3 &
``...The novelty of this work is a bit limited, as it combines existing techniques. ...'' (Negative) \\

Meta-review &
``...The paper studies an interesting problem and demonstrates strong empirical performance, although reviewers raised concerns regarding the methodological novelty of combining existing techniques. ...'' \\

\bottomrule
\end{tabular}
\caption{Case study illustrating conflict-aware aggregation under reviewer disagreement.}
\label{tab:case_study}
\end{table}

\section{Prompt Templates}
\label{appendix:prompts}

Our framework employs four prompt-based modules:
aspect-level opinion extraction
(Table~\ref{tab:prompt_opinion}),
opinion-level support factor evaluation
(Table~\ref{tab:prompt_support}),
review-level quality evaluation
(Table~\ref{tab:prompt_review_quality}),
and reliability-aware meta-review generation
(Table~\ref{tab:prompt_generation}).
For reproducibility, we present the corresponding prompt templates
used in each stage of the framework.

\begin{table*}[t]
\centering
\footnotesize
\setlength{\fboxsep}{6pt}
\renewcommand{\arraystretch}{0.95}

\fcolorbox{black!20}{gray!10}{
\parbox{0.92\textwidth}{

\setlength{\parskip}{0.2em}

You are an expert in extracting and structuring original sentences from academic paper reviews.

\vspace{0.2em}

Given a paper and its multiple reviews, perform the following tasks.

\vspace{0.2em}

Objectives:

(1) Extract clearly expressed opinion sentences from each individual review.

(2) Do not merge, deduplicate, summarize, or paraphrase opinions.

(3) Group the opinions according to review aspects.

(4) Annotate each opinion with its sentiment type.

(5) Strictly preserve the semantics of the original review text.

\vspace{0.2em}

Core Constraints:

(1) The opinion content must come directly from the review text.

(2) Prefer sentence-level extraction.

(3) Only minor formatting modifications are allowed.

(4) Do not explain the reviewer’s motivation.

(5) Do not rewrite sentences as ``the reviewer thinks/notes/says''.

(6) Do not merge multiple sentences into a new opinion.

(7) Do not introduce information not explicitly present in the review.

\vspace{0.2em}

Review Aspects (STRICT):

Each opinion must belong to one of the following six aspects only.

(1) novelty: Whether the paper presents new ideas, methods, or perspectives.

(2) soundness: Whether the technical approach, methodology, experiments, or reasoning are correct and reliable.

(3) clarity: Whether the paper is clearly written and easy to understand.

(4) advancement: Whether the work advances the state of the art or contributes meaningful progress to the field.

(5) compliance: Whether the paper follows conference requirements such as reproducibility, ethical standards, citations, or formatting.

(6) overall\_quality: The reviewer’s overall evaluation of the paper.

\vspace{0.2em}

Sentiment Types:

(1) positive: explicitly praises or affirms the paper.

(2) negative: explicitly criticizes or expresses dissatisfaction.

(3) neutral: objective description with no clear evaluative polarity.

\vspace{0.2em}

Special Rule:

If a sentence contains both positive and negative opinions, split it into two separate opinions whenever possible.

\vspace{0.2em}

Output Format:

The output must strictly follow JSON format.

\vspace{0.1em}

{\ttfamily
\footnotesize

\{

``paper\_id'': "<paper ID>",

``aspects'': \{

\hspace*{1em}``<aspect\_name>'': [

\hspace*{2em}\{

\hspace*{3em}``review'': "<reviewer ID>",

\hspace*{3em}``opinion'': "<original opinion sentence>",

\hspace*{3em}``sentiment'': "<positive | negative | neutral>"

\hspace*{2em}\}

\hspace*{1em}]

\}

\}

}

\vspace{0.2em}

Important Constraints:

(1) Output JSON only.

(2) Do not output explanations.

(3) Ensure the JSON can be parsed directly.

(4) Do not modify the original meaning of the sentence in \texttt{opinion}.

}
}

\caption{Prompt template for aspect-level opinion extraction.}
\label{tab:prompt_opinion}

\end{table*}

\begin{table*}[t]
\centering
\footnotesize
\setlength{\fboxsep}{6pt}
\renewcommand{\arraystretch}{0.95}

\fcolorbox{black!20}{gray!10}{
\parbox{0.92\textwidth}{

\setlength{\parskip}{0.2em}

You are an expert in academic peer review analysis.

\vspace{0.2em}

Given opinion sentences under the same review aspect, evaluate four support factors for each opinion.

\vspace{0.2em}

Each factor should be rated on a scale from 1 to 5:

(1) 1 = very weak / not satisfied at all

(2) 2 = weak

(3) 3 = moderate

(4) 4 = strong

(5) 5 = very strong

\vspace{0.2em}

Important Constraints:

(1) Use the full range (1--5); do not collapse everything to 3 or 4.

(2) Be discriminative: different opinions should receive different scores whenever appropriate.

(3) Avoid assigning identical scores unless the opinions are truly similar in quality.

\vspace{0.2em}

Factor Definitions:

(1) Specificity:
Does the opinion refer to concrete components of the paper, such as methods, experiments, datasets, or results?

(2) Rationality:
Is the opinion logically sound and non-emotional?

(3) Justification:
Does the opinion provide reasoning, explanation, or supporting evidence?

(4) Academic Plausibility:
Is the opinion a plausible and meaningful academic judgment?

\vspace{0.2em}

Output Format:

The output must strictly follow JSON format.

\vspace{0.1em}

{\ttfamily
\footnotesize

\{

``opinions'': [

\hspace*{1em}\{

\hspace*{2em}``review'': "<reviewer ID>",

\hspace*{2em}``opinion'': "",

\hspace*{2em}``sentiment'': "",

\hspace*{2em}``f1\_specificity'': 1,

\hspace*{2em}``f2\_rationality'': 1,

\hspace*{2em}``f3\_justification'': 1,

\hspace*{2em}``f4\_academic\_plausibility'': 1

\hspace*{1em}\}

]

\}

}

\vspace{0.2em}

Important Constraints:

(1) Output JSON only.

(2) Do not output explanations.

(3) Ensure the JSON can be parsed directly.

(4) Assign integer scores only.

}
}

\caption{Prompt template for opinion-level support factor evaluation.}
\label{tab:prompt_support}

\end{table*}

\begin{table*}[t]
\centering
\footnotesize
\setlength{\fboxsep}{6pt}
\renewcommand{\arraystretch}{0.95}

\fcolorbox{black!20}{gray!10}{
\parbox{0.92\textwidth}{

\setlength{\parskip}{0.2em}

You are an expert in evaluating the quality of academic peer reviews.

\vspace{0.2em}

You will be given all opinion sentences written by a single reviewer for a specific paper. These sentences have already been extracted from the review text and classified by review aspect and sentiment.

\vspace{0.2em}

Your task is not to judge whether the reviewer is correct. Instead, evaluate the overall review quality across four dimensions.

\vspace{0.2em}

Evaluation Dimensions:

(1) clarity:
Are the comments easy to understand and clearly written?

(2) specificity:
Do the comments reference concrete aspects of the paper?

(3) constructiveness:
Do the comments provide useful suggestions or actionable feedback?

(4) professionalism:
Are the comments respectful, objective, and professional?

\vspace{0.2em}

Scoring Rules:

Each dimension must be scored from 1 to 5.

(1) 1 = very poor

(2) 2 = weak

(3) 3 = moderate

(4) 4 = good

(5) 5 = excellent

\vspace{0.2em}

Important Constraints:

(1) Each opinion is associated with a weight between 0 and 1, indicating its reliability.

(2) Opinions with lower weights are less trustworthy and should have less influence on the final judgment.

(3) Focus more on high-weight opinions when evaluating overall review quality.

(4) Provide concise reasons for each quality score.

\vspace{0.2em}

Output Format:

The output must strictly follow JSON format.

\vspace{0.1em}

{\ttfamily
\footnotesize

\{

``review'': "<reviewer ID>",

``quality\_scores'': \{

\hspace*{1em}``clarity'': \{

\hspace*{2em}``score'': 1,

\hspace*{2em}``reason'': "..."

\hspace*{1em}\},

\vspace{0.1em}

\hspace*{1em}``specificity'': \{

\hspace*{2em}``score'': 1,

\hspace*{2em}``reason'': "..."

\hspace*{1em}\},

\vspace{0.1em}

\hspace*{1em}``constructiveness'': \{

\hspace*{2em}``score'': 1,

\hspace*{2em}``reason'': "..."

\hspace*{1em}\},

\vspace{0.1em}

\hspace*{1em}``professionalism'': \{

\hspace*{2em}``score'': 1,

\hspace*{2em}``reason'': "..."

\hspace*{1em}\}

\},

\vspace{0.1em}

``overall\_assessment'': "brief summary"

\}

}

\vspace{0.2em}

Important Constraints:

(1) Output JSON only.

(2) Do not output explanations outside the JSON object.

(3) Ensure the JSON can be parsed directly.

(4) All scores must be integers between 1 and 5.

}
}

\caption{Prompt template for review-level quality evaluation.}
\label{tab:prompt_review_quality}

\end{table*}

\begin{table*}[t]
\centering
\footnotesize
\setlength{\fboxsep}{6pt}
\renewcommand{\arraystretch}{0.95}

\fcolorbox{black!20}{gray!10}{
\parbox{0.92\textwidth}{

\setlength{\parskip}{0.2em}

You are an experienced Area Chair writing a meta-review for a top-tier academic conference.

\vspace{0.2em}

You are given a structured evidence package extracted from peer reviews.

\vspace{0.2em}

The following processing steps have already been applied:

(1) sentence-level opinion extraction

(2) aspect grouping

(3) conflict detection

(4) sentence-level reliability estimation

(5) reviewer-level reliability estimation

\vspace{0.2em}

Each opinion have one reliability weight:
final\_weight

\vspace{0.2em}

Evidence Interpretation Rules:

(1) Give higher attention to opinions with larger \texttt{final\_weight}.

(2) Opinions with lower \texttt{final\_weight} should have minimal influence.

(3) When conflicting opinions exist, weigh them according to \texttt{final\_weight}.

(4) Do not quote reviewer sentences verbatim.

\vspace{0.2em}

Writing Guidelines:

(1) Write one to three coherent paragraphs.

(2) Maintain a professional academic tone.

(3) Summarize major strengths and weaknesses.

(4) Discuss reviewer disagreements when necessary.

(5) Provide a final recommendation from:
accept / weak accept / borderline / weak reject / reject.

\vspace{0.2em}

Constraints:

(1) Do not mention reviewers explicitly.

(2) Do not output lists or bullet points.

(3) Do not introduce information not supported by the evidence.

\vspace{0.2em}

Output Format:

Output only the meta-review text without additional explanations or formatting.

\vspace{0.2em}

Structured Evidence:

\texttt{\{structured\_evidence\}}

}
}

\caption{Prompt template for reliability-aware meta-review generation.}
\label{tab:prompt_generation}

\end{table*}

\end{document}